\documentclass{article}
\usepackage{graphicx}

\usepackage{times}
\usepackage{helvet}
\usepackage{courier}

\newcommand{\eq}[1]{Eq.~(\ref{eq.#1})} 

\newcommand{\fig}[1]{Fig.~\ref{fig.#1}}

\newcommand{\tbl}[1]{Table~\ref{table.#1}}

\newcommand{\eqlabel}[1]{\label{eq.#1}}
\newcommand{\figlabel}[1]{\label{fig.#1}}
\newcommand{\tbllabel}[1]{\label{table.#1}}

\newcommand{\Flux}{{\bf F}}
\newcommand{\Reynolds}{\mbox{Re}}
\newcommand{\sensorSize}{\ell}
\newcommand{\sourceLength}{\lambda}

\newcommand{\density}{\ensuremath{\rho}}
\newcommand{\viscosity}{\ensuremath{\eta}}
\newcommand{\vFluid}{{\bf v}}
\newcommand{\vCell}{v_{\mbox{\scriptsize cell}}} 

\newcommand{\Dsmall}{D_{\mbox{\scriptsize small}}}
\newcommand{\Dlarge}{D_{\mbox{\scriptsize large}}}

\newcommand{\meter}{\mbox{m}}
\newcommand{\millimeter}{\mbox{mm}}
\newcommand{\micron}{\mbox{$\mu$m}}
\newcommand{\second}{\mbox{s}}
\newcommand{\molecule}{\mbox{molecule}}
\newcommand{\kg}{\mbox{kg}}
\newcommand{\gram}{\mbox{g}}

\newcommand{\Pascal}{\mbox{Pa}}

\begin{document}

\title{Modeling Microscopic Chemical Sensors in Capillaries}
\author{Tad Hogg\\Hewlett-Packard Laboratories\\Palo Alto, CA}

\maketitle

\begin{abstract}
Nanotechnology-based microscopic robots could provide accurate
\textit{in vivo} measurement of chemicals in the bloodstream for
detailed biological research and as an aid to medical treatment.
Quantitative performance estimates of such devices require models of
how chemicals in the blood diffuse to the devices. This paper models
microscopic robots and red blood cells (erythrocytes) in capillaries
using realistic distorted cell shapes. The models evaluate two
sensing scenarios: robots moving with the cells past a chemical
source on the vessel wall, and robots attached to the wall for
longer-term chemical monitoring. Using axial symmetric geometry with
realistic flow speeds and diffusion coefficients, we compare
detection performance with a simpler model that does not include the
cells. The average chemical absorption is quantitatively similar in
both models, indicating the simpler model is an adequate design
guide to sensor performance in capillaries. However, determining the
variation in forces and absorption as cells move requires the full
model.
\end{abstract}

\section{Introduction}

Nanotechnology has the potential to greatly improve health
care~\cite{morris01,nih03,keszler01}. For example, nanoscale
particles can significantly enhance medical imaging~\cite{vodinh06}
and drug delivery~\cite{allen04}. Future possibilities include
programmable machines comparable in size to cells and able to sense
and modify their environments. Such microscopic robots could provide
significant medical benefits~\cite{freitas99,morris01} with
precisely controlled targeted actions at the scale of individual
cells.

While microscopic robots cannot yet be manufactured, current
laboratory demonstrations indicate their likely minimal capabilities
based on nanoscale electronics, sensors and
motors~\cite{barreiro08,berna05,collier99,craighead00,howard97,fritz00,montemagno99,soong00,wang05}.
These capabilities and estimates of the properties of \textit{in
vivo} environments where the devices would operate are sufficient to
develop computational models of device performance for a variety of
medical tasks.
These devices utilize stronger materials and faster computation than
is possible with biologically-based
machines~\cite{weiss06,benenson04,win08}.

A particularly important robot environment is the blood vasculature,
which allows the robots ready access throughout the body via passive
movement with the fluid. A challenge for computational models of
devices in blood is the interactions among the blood components and
the devices. The blood is not a uniform fluid on the scale of the
devices but rather consists of larger objects (the cells) embedded
in a fluid (the plasma). Thus detailed models must include fluid
dynamics with cells that move and distort, branching vessels,
changes in vessel size, and hydrodynamic interactions among the
devices. Of particular interest is the behavior of microscopic
devices in capillaries, which exchange chemicals with body tissues
and bring the devices close to cells throughout the body.

Detailed models including numerous cells interacting with the
devices are computationally intensive. Thus design studies of
medical task scenarios benefit from simpler models that nevertheless
give reasonable quantitative estimates of task performance. One such
model considers the blood as a homogeneous fluid, with possibly
modified parameters, at the scale of the robots. One important
process to model is transport of chemicals to the devices, which is
similar to the process for bacteria~\cite{berg77}. Chemical
transport is key for medical tasks such as high-resolution
diagnosis~\cite{hogg06b}, treatment through targeted drug
delivery~\cite{allen04,li06,freitas06,tucker08}, microsurgery within
and among individual cells~\cite{freitas99,sretavan05,hogg05} and
power generation using chemicals available in the bloodstream. An
example of power generation in the robots is combining glucose and
oxygen in fuel cells analogous to the enzyme-mediated reactions in
bacteria~\cite{chaudhuri03}.

One modeling approximation, suited to vessels wide enough to
accommodate multiple cells across their diameter, treats the effect
of cells as increasing diffusion in vessels large enough to allow
cells to rotate~\cite{keller71}. In this case, a model with
homogeneous fluid and an increase in the chemical diffusion
coefficients gives a reasonable approximation to chemical transport
in the vessel. However, in capillaries cells move through single
file so this approach is not suitable.

As an aid to modeling behavior in capillaries, this paper compares a
simple model (homogeneous fluid with no cells) with the effect of
cells moving in the fluid. The evaluation of behavior with cells is
simplified to use realistic but approximate geometry for deformed
cell shape based on prior numerical studies of mechanical properties
of cells and their membranes. This approach avoids the computational
cost of simultaneously solving for the shape of the cell due to
interaction with fluid, while still being useful to calibrate models
not including cells. The models considered here are similar to
previous studies of chemical transport in blood
vessels~\cite{popel89}, but with scenarios relevant for microscopic
robots. In particular, the robots considered here, of size
comparable to bacteria, are somewhat smaller than blood and tissue
cells, and so can involve steeper concentration gradients. In
particular, this model captures two key effects of cells in
capillaries: their effect on the fluid flow and their restriction on
chemical diffusion within the plasma. The remainder of the paper
describes the cell model and compares its predictions with the
simpler homogeneous fluid model.

\section{Method}

We examine the effect of cells on microscopic chemical sensors in
capillaries by numerically evaluating the fluid flow and chemical
diffusion. We consider a cylindrical vessel~\cite{krogh19} and an
axial-symmetric geometry with the cells centered along the axis of
the vessel. Fluid motion in these small vessels is dominated by
viscosity: the Reynolds number of the flow, i.e., ratio of inertial
to viscous forces, is small: $\Reynolds \equiv \density v
\ell/\viscosity \ll 1$, where $\density$ and $\viscosity$ are the
fluid density and viscosity, respectively, $v$ and $\ell$ are
characteristic speed of the flow and size of the vessel and objects
in the fluid. Such microfluidic flows are generally smooth and
laminar with little mixing~\cite{squires05}, leading to physical
behaviors distinct from those familiar with macroscopic
flows~\cite{purcell77,vogel94}. This creeping or Stokes flow is
relatively simple, with the computational challenge arising from the
moving objects in the flow rather than from the nonlinearities in
the fluid flow itself~\cite{fauci01}.

\begin{table}
\begin{center}
\begin{tabular}{|l|l|}
  \hline
  \textbf{parameter} & \textbf{value} \\ \hline
  vessel radius & $R=3\,\micron$ \\
  fluid density & $\density=10^3\,\kg/\meter^3$ \\
  fluid viscosity & $\viscosity=10^{-3}\,\Pascal \cdot \second$ \\
  hematocrit & $h=25\%$ \\
  cell speed & $\vCell=0.2, 1 \mbox{ and } 2\,\millimeter/\second$ \\
  \hline \multicolumn{2}{|c|}{cell geometry}\\
  cell volume & $V=90\,\micron^3$ \\
  cell surface area & $S=135\,\micron^2$ \\
  cell spacing & $L=V/(\pi R^2 h) = 12.7\,\micron$ \\
  \hline \multicolumn{2}{|c|}{diffusion coefficients}\\
  small molecules & $\Dsmall = 2\times 10^{-9}\,\meter^2/\second$\\
  large molecules & $\Dlarge = 10^{-10}\,\meter^2/\second$\\
  \hline
\end{tabular}
\end{center}
\caption{\tbllabel{parameters}Model parameters. The volume and
surface area are typical values for red blood cells~\cite{secomb01}.
The vessel size and hematocrit (fraction of vessel volume occupied
by cells) are typical for capillaries, and fluid density and
viscosity for plasma are similar to the values of water at body
temperature~\cite{freitas99}.}
\end{table}

Applications for microscopic devices involve both small and large
molecules dissolved in the fluid. As examples, oxygen may be used in
a fuel cell to power the device and large molecules, such as
signaling proteins, are significant for diagnostics. The diffusion
coefficients for such molecules in fluids at body temperature range
from $2\times 10^{-9}\,\meter^2/\second$ for small molecules (e.g.,
oxygen and carbon dioxide) to around $10^{-10}\,\meter^2/\second$
for a typical 10-kilodalton protein~\cite{freitas99}. We examine
behavior for both these diffusion coefficients and a range of
typical flow speeds in capillaries.
\tbl{parameters} gives the model parameters for the vessel, fluid
properties, cell geometry and chemical diffusion.

\subsection{Cell Geometry}

Unlike the biconcave shape of red blood cells at rest, cells moving
through narrow vessels are considerably
distorted~\cite{secomb01,pozrikidis05}. Nevertheless, the cells
maintain their volume and surface area since their contents are
nearly incompressible and the membrane strongly resists changes in
area~\cite{pries96}. More complex interactions between cells occur
where vessels bend or branch~\cite{boryczko03}.

Significantly for chemical transport through the plasma to sensors,
the flow establishes a gap between the distorted cell and the vessel
wall. The gap between the cell and vessel wall depends on the speed
of the cell in the vessel, as indicated in \tbl{gap size} for a
vessel of radius $R=3\,\micron$. As flow speed increases, the gap
increases and cells become narrower and longer.

\begin{table}
\begin{center}
\begin{tabular}{|l|ccc|}
  \hline
  cell speed ($\millimeter/\second$)& $0.2$ & $1$ & $2$ \\ \hline
  gap ($\micron$) & $0.7$ & $0.9$ & $1.0$ \\
  \hline
\end{tabular}
\end{center}
\caption{Gap between cell and vessel wall as a function of cell
speed, $\vCell$, in a capillary with radius of three
microns~\cite{secomb01}.\tbllabel{gap size}}
\end{table}

To estimate the effect on chemicals reaching microscopic sensors, we
consider an approximate geometry for the cells including most of the
distortion from the flow. For a vessel of uniform diameter far from
its merge with other vessels, we consider the cells as achieving
their equilibrium shape with respect to the flow and avoid
explicitly modeling the forces distorting the cell. With this
geometry, we then treat the cells as rigid (i.e., assuming the
residual forces on the cells are small enough to not give
significant further distortion), thereby considerably simplifying
the subsequent numerical evaluation of fluid flow and chemical
diffusion.

We describe each cell as a solid of revolution defined by the curve
illustrated in \fig{cell geometry}. This curve consists of parts of
three circles and a line segment. The curve is continuous and
smooth, i.e., the parts match with continuous first derivative.
Specifically, the front of the cell model consists of a
quarter-circle, with radius $r$, connected to a straight line
parallel to the vessel axis of length $a$. The back of this line
connects to a small half-circle which connects to another
quarter-circle, of radius $s$, forming the trailing edge of the cell
model. The full 3-dimensional form of the cell consists of rotating
the area enclosed by these curves around the vessel axis, giving an
axially symmetric cell shape. This geometry, similar to that
employed in some models of oxygen transport~\cite{vadapalli02},
approximates shapes obtained through numerical evaluation of the
fluid forces on the cells~\cite{secomb01,pozrikidis05}.

\begin{figure}
\begin{center}
    \includegraphics{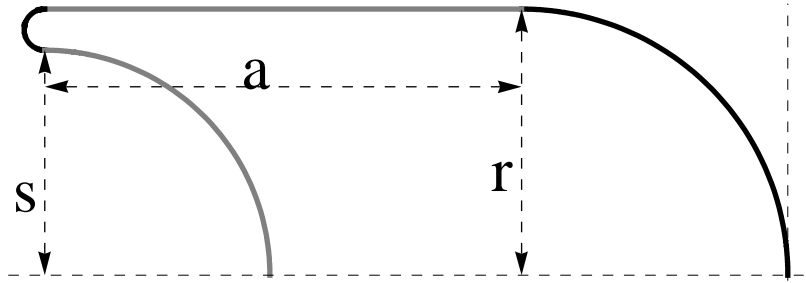}
\end{center}
\caption{\figlabel{cell geometry}Cell geometry showing a cross
section of half the vessel, from the axis on the bottom to the
vessel wall at the top. The model involves 4 joined curves, shown as
alternating black and gray parts. The cell moves from left to right
through the vessel.}
\end{figure}

We determine the three geometric parameters of the cell model, $r$,
$a$ and $s$, from three constraints. First, we pick $R-r$ to equal
the gap between the cell and vessel wall given in \tbl{gap size},
where $R$ is the vessel radius. Second, we set $a$ and $s$ to match
the cell volume and surface area for the 3-dimensional shape given
in \tbl{parameters}. \fig{geometries} shows the resulting cell model
shapes for various speeds of the cells through the vessel.

\begin{figure}
\begin{center}
\includegraphics{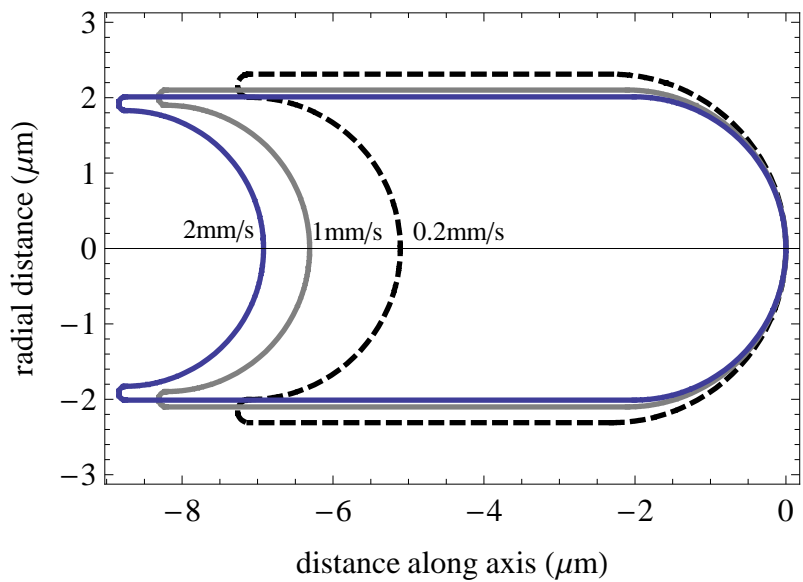}
\end{center}
\caption{\figlabel{geometries}Cell geometries for speeds
$\vCell=0.2$, $1$ and $2\millimeter/\second$. The figure shows a
cross section through the vessel, centered on the vessel axis. The
front of each cell is shown in the same location along the vessel,
at the right. The cell and fluid move from left to right.}
\end{figure}

Cells pass in single file through the capillary. For simplicity, we
consider the cells to all have the same size and to be uniformly
spaced. The spacing between cells is determined by the hematocrit
value, $h$, in capillaries, i.e., the fraction of volume occupied by
red blood cells. The hematocrit in capillaries is generally somewhat
smaller than in larger vessels.
With distance $L$ between the fronts of successive cells, as shown
in \fig{spacing}, the cells occupy a fraction $V/(\pi R^2 L)$ of the
vessel volume, which corresponds to the hematocrit $h$. Equating
these values gives the value of $L$ in \tbl{parameters}.

\begin{figure}
\begin{center}
\includegraphics{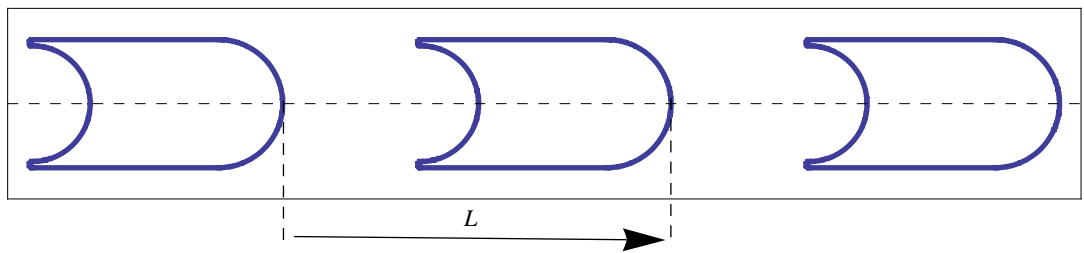}
\end{center}
\caption{\figlabel{spacing}Cell spacing, using the geometry for
speed of $1\millimeter/\second$. The figure shows a cross section
through the vessel, centered on the vessel axis. The distance
between the front of successive cells is $L$. Cells and fluid move
from left to right through the vessel. The upper and lower lines
represent the vessel wall, while the left and right lines are inlet
and outlet boundaries, respectively, for a finite section of a
conceptually infinitely long vessel.}
\end{figure}

\subsection{Chemical Sensor}

\begin{table}
\begin{center}
\begin{tabular}{|l|l|}
  \hline
  \textbf{parameter} & \textbf{value} \\ \hline
  sensor length & $\sensorSize=2\,\micron$ \\
  \hline \multicolumn{2}{|c|}{scenario 1: sensor band on vessel wall}\\
  chemical inflow concentration & $C=10^{17}\,\molecule/\meter^3$ \\
  \hline \multicolumn{2}{|c|}{scenario 2: sensor moving with fluid}\\
  chemical source length & $\sourceLength=10\,\micron$ \\
  chemical source flux & $K=10^{13}\molecule/\second/\meter^2$ \\
  \hline
\end{tabular}
\end{center}
\caption{\tbllabel{sensor scenarios}Parameters for two sensing
scenarios. The sensor size is a design choice commonly considered in
studies of microscopic sensors~\cite{freitas99,hogg06a}, and small
enough to pass through even the smallest blood vessels. The source
length considered here, for a chemical source on the vessel wall, is
comparable to the size of a single cell which provides a challenging
detection task for microscopic sensors. The choice of inlet
concentration and source flux are discussed in the text.}
\end{table}

For the chemical sensor, we consider two scenarios, with parameters
given in \tbl{sensor scenarios}. To highlight the difference between
a simple model with no cells and a more accurate model containing
the cells, we consider the chemical as existing in the plasma only,
and not passing into or out of the cells. Chemicals passing through
the cell membranes, e.g., oxygen, give less difference between the
two modeling approaches.

In the first scenario, appropriate for long term monitoring, the
sensor is a group of robots forming a uniform band embedded in the
vessel wall. This band geometry maintains axial symmetry. In this
scenario, the chemical of interest is in the plasma surrounding the
cells in the vessel and is brought past the sensor by the blood
flow. The main performance question is how much of the chemical in
the fluid reaches the sensor for capture rather than being driven
past the sensor by the fluid flow.

The band of sensors around the vessel wall could consist of multiple
rings of devices, but for simplicity we consider the most
challenging case for chemical detection: the smallest possible size
consistent with axial symmetry of a single device length along the
length of the vessel, i.e., the band has length $\sensorSize$ along
the vessel and consists of about $2\pi R/\sensorSize \approx 9$
individual sensors forming the band around the vessel. The sensor
band has a total surface area of $2 \pi R \sensorSize =
38\micron^2$.

Such a sensor band could most readily be constructed from a
collection of microscopic devices moving through the circulation and
aggregating at a suitable location, e.g., marked by a pattern of
chemicals of interest~\cite{hogg06a}. In this way, individual
devices could be small enough to flow through even the smallest
capillaries and build a larger aggregate structure on the vessel
wall. The process whereby such devices link to each other to form
such a structure could employ simple local rules similar to those
proposed for larger-scale reconfigurable
robots~\cite{fukuda88,bennett00,bojinov02,arbuckle04,salemi01}.

In the second sensing scenario, we consider a single sensor flowing
in the vessel between two cells and the chemical source is a single
cell-sized region on the vessel wall. In this case we take the
sensor to be spherical with diameter $\sensorSize$, and moving in
the center of the vessel, thereby maintaining axial symmetry. The
specific choice of positioning of the sensor, half-way between the
two cells, can be viewed either as an approximation of where it
would be moved by the fluid, or as the device having active
locomotion to maintain this position with respect to the cells. This
model assumes the presence of the sensor makes negligible changes in
the shape or positioning of the nearby cells.

This scenario corresponds to robots patrolling the circulation for
chemical events of interest. Thus the performance measure is not
only how much chemical the sensor detects but also how much it
detects while close to the source. That is, the patrolling function
can involve not only detecting the existence of a chemical source
but also locating it for possible subsequent operations, such as
moving to the source location~\cite{hogg06a} and releasing drugs at
this location~\cite{freitas06,li06}. To maintain axial symmetry, we
take the source region to be a band around the vessel of length
$\sourceLength=10\micron$, i.e., about as large as a single cell
wrapped around the vessel, thereby testing how quickly the passing
sensor can detect the fairly steep concentration gradient produced
by chemicals released by a single cell. We take the chemical flux to
be uniform throughout this band.

For specific numerical values of the rate at which the sensors
absorb molecules we need values for the inlet concentration $C$ and
source flux $K$ in the first and second scenarios, respectively. The
concentration throughout the vessel and flux to the sensor are
linear in these values. Thus the relative comparison between models
with and without cells does not depend on the values of $C$ and $K$.
Nevertheless, for definiteness, we choose specific values, given in
\tbl{sensor scenarios}, corresponding to a typical high-resolution
diagnostic task. Other situations, with either higher or lower
concentrations, simply proportionally change the sensor flux values
reported below.

The amount of a chemical in a milliliter of blood plasma relevant
for medical diagnostics range from picograms for some proteins to
milligrams (e.g., for glucose)~\cite{freitas99,service08}. High
resolution diagnostics for which microscopic robots would be useful
involve the lower concentrations, in particular with robots passing
close to the sources of rare chemicals where the local
concentrations will be considerably larger than these measured
values in blood samples, which correspond to the chemical diluted
throughout the blood volume. For example, suppose a small source
released a chemical into a few nearby capillaries giving local
concentration $C$ near the source. After mixing throughout the full
$5$-liter blood volume of an adult, a concentration $C$ in a typical
capillary, of length about a millimeter, dilutes by about a factor
of $10^{11}$. As a definite value for the first scenario, we
consider a 10-kilodalton protein with concentration $2\times
10^{-9}\,\gram/\mbox{cm}^3$, which corresponds to
$C=10^{17}\,\molecule/\meter^3$.
A corresponding flux from a single cell producing this concentration
in a nearby capillary is around
$K=10^{13}\,\molecule/\second/\meter^2$, which gives a challenging
diagnostic task~\cite{hogg06a}. For a cell with surface area
$300\micron^2$, this value corresponds to producing
$3000\,\molecule/\second$, or $5\times 10^{-17}\,\gram/\second$,
through the cell membrane.
We use these example values for the two scenarios, summarized in
\tbl{sensor scenarios}.

\subsection{Numerical Method}

With the specified geometry of cells and sensors, we solve the
partial differential equations for fluid flow and chemical transport
numerically using the finite element method~\cite{web.comsol}. For
numerical solution we model a segment of the vessel, as illustrated
in \fig{spacing} except using 10 and 20 cells in the models for the
first and second scenario, respectively. In the first scenario, the
model also includes the spherical sensor half-way between the middle
two cells.

The Navier-Stokes equation governs the fluid
flow~\cite{karniadakis05}. For a periodic array of cells shown in
\fig{spacing} we specify the speed $\vCell$ at which the cells move
through the vessel. A convenient numerical method is using a frame
of reference moving with the cells~\cite{mauroy07,vadapalli02},
analogous to the view from a camera moving at speed $\vCell$. In
this moving reference frame, the cells are stationary while the
vessel wall moves backwards. This means the fluid flow does not
change with time and we compute a steady-state solution for the flow
throughout the vessel in the moving reference frame.

The fluid boundary conditions are zero velocity (``no slip'') on the
boundaries of the cells and moving backwards with speed $-\vCell$
along the vessel wall. As part of the numerical solution, we find
the pressure difference giving no net force on the cell in the
center of the modeled vessel region. This pressure difference
corresponds to the fluid pushing the cells through the vessel with a
steady speed $\vCell$.

For chemicals in the fluid, the concentration $c$ is governed by the
diffusion equation~\cite{berg93}
\begin{equation}\eqlabel{diffusion}
\frac{\partial c}{\partial t} = -\nabla \cdot \Flux
\end{equation}
where $\Flux = -D \nabla c + \vFluid c$ is the chemical flux, i.e.,
the rate at which molecules pass through a unit area, and $\vFluid$
is the fluid velocity vector determined from the solution of the
Navier-Stokes equation. The first term in the flux is diffusion,
which acts to reduce concentration gradients, and the second term is
motion of the chemical due to the movement of the fluid in which the
chemical is dissolved. Chemical reactions that create or destroy the
chemical in the fluid give additional terms in the diffusion
equation.

As boundary conditions for the diffusion equation, we take the cells
and vessel wall to be insulating, i.e., no chemical crosses those
boundaries. Sufficiently far from the sensor, diffusion makes the
concentration uniform throughout the vessel. In our case, far
upstream from the sensor we have a known concentration at the inlet
to the modeled vessel segment, described below for the two
scenarios. Far downstream, the concentration is again uniform but
its value depends on how much of the chemical is absorbed by the
sensor. Thus the downstream concentration is not known a priori, and
we use instead a convective boundary condition at the outlet, i.e.,
taking $\nabla c = 0$, corresponding to a uniform concentration
throughout the vessel far downstream of the sensor.

The boundary conditions for the vessel inlet and the sensor depend
on the scenario, as does the initial condition for concentration in
the vessel. A natural boundary condition for the sensor is zero
concentration, corresponding to a completely absorbing sensor. This
gives the maximum rate the sensor could detect chemicals and is
suitable for low concentration conditions where the sensor is
diffusion-limited~\cite{sheehan05}. This will generally be the
situation for applications: detecting chemicals with low
concentrations and localized in small volumes which cannot be
readily detected in simpler laboratory procedures, e.g., from a
blood sample. Alternatively, the sensor could have active pumps to
collect chemicals~\cite{freitas99}, thereby maintaining near-zero
concentrations even at higher flux rates.

In the first scenario, we suppose a chemical is released into the
vessel well upstream of the sensor band. Eventually the fluid brings
the chemical past the sensor band, which absorbs some of the
chemical. We model this situation as initially zero concentration in
the vessel and a constant concentration $C$ at the inlet.

Instead of a zero concentration boundary condition for the sensor,
with the sensor band moving backwards with the wall and a zero flux
condition on the rest of the wall, a simpler numerical approach is a
time-dependent flux boundary condition for the entire wall. That is,
at time $t$, when the sensor has moved a distance $-\vCell t$ from
its initial location on the wall, we take flux to be $k c$ along the
sensor and zero elsewhere on the wall, smoothed over a short
distance, $0.2\,\micron$, to avoid discontinuity in the numerical
solution. We pick $k$ large enough so the concentration along the
sensor is very small, but not so large as to introduce numerical
instabilities. In our case, $k=1\meter/\second$ is sufficient to
make the concentration at the sensor less than $10^{-4}C$ where $C$
is the inlet concentration. Thus this procedure approximately models
a fully absorbing sensor.
We start with zero concentration in the vessel and solve in time
until the flux reaching the vessel outlet stabilizes. At this point
the flux reaching the sensor becomes periodic in time, with no
change after a shift of one cell spacing, i.e., distance $L$, past
the sensor.

In the reference frame fixed on the cells, the sensor band moves
backwards at speed $\vCell$ which would eventually cause the band to
move outside the segment of the vessel included in the model. To
allow solving over long times, we can exploit the periodic nature of
the fluid flow. Specifically, we solve incrementally over a time
$L/\vCell$ where a given location on the wall moves past one cell
and then shift the solution by a distance $L$ to place the absorber
back in the center of the solution domain and repeat. This procedure
keeps the region of interest on the wall, i.e., the absorbing band,
near the center of the model domain.

In the second scenario, we take the chemical source band on the
vessel wall to produce the chemical with a uniform flux $K$. We
solve this model in the reference frame moving with the cells, so
the source on the wall moves backwards past the cells and sensor. As
described above, we model this with a time-dependent boundary
condition with flux $K$ at the location of the source and zero
elsewhere on the vessel wall, smoothed over a short distance,
$0.2\,\micron$. The inlet has zero concentration.

For the initial concentration, when the source is well downstream of
the absorbing sensor, all the chemical released by the source is
swept downstream. Thus the concentration upstream of the source is
zero and downstream the flow must have the entire production of the
source. The source produces chemical at a rate $2 \pi R
\sourceLength K$. If the concentration far downstream is $C$, all of
which is in the plasma, which occupies volume fraction $1-h$ of the
vessel, then the fluid moves chemical at a rate $C \pi R^2 \vCell
(1-h)$. Equating these expressions gives the initial downstream
concentration of $C = 2 \sourceLength K / (R \vCell (1-h))$. In the
simpler model without cells, the plasma fills the whole vessel so
$h=0$ in this expression for the downstream concentration in that
case. As initial condition, we take a smooth transition, over a
distance of $10\micron$, between 0 and this value of $C$ at the
initial location of the source. We start the numerical solution with
the source well downstream of the sensor, so the precise form of
this initial concentration has little effect on the behavior of the
sensor by the time it passes the source.

\section{Results}

We compare the sensor performance in the two scenarios for the
models with and without cells using the parameters of
\tbl{parameters}. The numerical solutions give the rate the sensors
absorb the chemical. For the evenly spaced cells with identical
geometry considered here, the fluid flow is independent of time and
periodic, with period $L$, along the length of the vessel. In the
reference frame moving with the cells, the fluid circulates between
the cells, moving away from the front of the cell along the vessel
axis and toward the front of the cell near the vessel wall.

\subsection{Sensor Band on Vessel Wall}

In the first scenario, the chemical continually enters the vessel
inlet and some reaches the sensor on the wall. The flux to the
sensor is nearly periodic in time: in each time interval $L/\vCell$
the cells move along the vessel by one cell spacing, i.e, distance
$L$, relative to the sensor band. After this shift, the geometry,
fluid flow, concentration and hence the flux to the sensor are the
same as before.

\begin{figure}
\begin{center}
\includegraphics*[viewport=0 90 216 135]{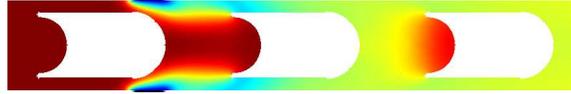}
\end{center}
\caption{\textbf{Sensor on the vessel wall.} Concentration around
three cells passing the sensor band on the vessel wall. Geometry is
as in \fig{spacing}. Concentration ranges from 0 (dark blue) to
$10^{17}\,\molecule/\meter^3$ (red). The black lines along the
vessel wall mark the location of the sensor at this time, with
length $\sensorSize=2\micron$. The cells and fluid move from left to
right. Cell speed is $\vCell=1\millimeter/\second$ and diffusion
coefficient is $\Dlarge$.}\figlabel{concentration band}
\end{figure}

\begin{table}
\begin{center}
\begin{tabular}{|l|ccc|}
  \hline
  cell speed ($\millimeter/\second$)& $0.2$ & $1$ & $2$ \\
  pressure gradient ($\Pascal/\meter$) & $1.8\times 10^5$ & $9.1\times 10^5$ & $1.8\times 10^6$ \\
 \hline maximum force on sensor band (pN) & $10.2$ & $51$ & $102$ \\
 $\ldots$ without cells & $10.0$ & $50$ & $100$ \\
  \hline \multicolumn{4}{|c|}{large molecules}\\
  average count rate ($\second^{-1}$) & $260$ & $570$ & $800$ \\
  $\ldots$ without cells & $310$ & $590$ & $760$ \\
  \hline \multicolumn{4}{|c|}{small molecules}\\
  average count rate ($\second^{-1}$) & $490$ & $1700$ & $3700$ \\
  $\ldots$ without cells & $560$ & $2500$ & $4200$ \\
  \hline
\end{tabular}
\end{center}
\caption{Behavior of the sensor band on the vessel
wall.}\tbllabel{results: band}
\end{table}

\fig{concentration band} shows how the sensor band absorbs the
chemical primarily from the layer of fluid near the wall.
Comparing the different flow speeds and diffusion coefficients in
\tbl{results: band} shows the sensor captures more chemical at
faster speeds, though a decreasing fraction of the total inlet flux,
which grows linearly with the fluid speed. The values in the two
models, with and without cells, are comparable, indicating the
simpler ``empty vessel'' model is a reasonable approximation.

For comparison, an absorbing sphere of radius $r$ in a region with a
chemical with diffusion coefficient $D$ and concentration $C$ far
from the sphere, absorbs the chemical at a rate~\cite{berg93}
\begin{equation}\eqlabel{absorb}
4 \pi D a C
\end{equation}
Thus a sphere whose surface area is the same as that of the sensor
band, i.e., $2 \pi R \sensorSize$, absorbs at a rate
$200\,\molecule/\second$ when the chemical has diffusion coefficient
$\Dlarge$ and concentration $C$ of \tbl{sensor scenarios}.

For sensors attached to the vessel wall, an important question is
how much force they require to remain attached to the wall.
\fig{force on band} shows how the shear force on the sensor band
varies as cells pass, and compares with the force in the
corresponding empty vessel model. In the empty vessel, the flow in a
pipe of radius $R$ gives shear force per unit area on the wall of $4
v \viscosity/R$ where $v$ is the average speed along the vessel,
which we take equal to $\vCell$ for this comparison. Multiplying by
the surface area of the sensor band, $2 \pi R \sensorSize$, gives
the force on the sensor band in the empty vessel~\cite{king02}.
We see the forces in the two models are comparable, so the empty
vessel model gives a useful guideline. Moreover, as seen in
\fig{force on band} the force varies noticeably, by about $25\%$, as
the cell passes, which could be detected with sensors for fluid
shear~\cite{ghosh03}.

\begin{figure}
\begin{center}
\includegraphics{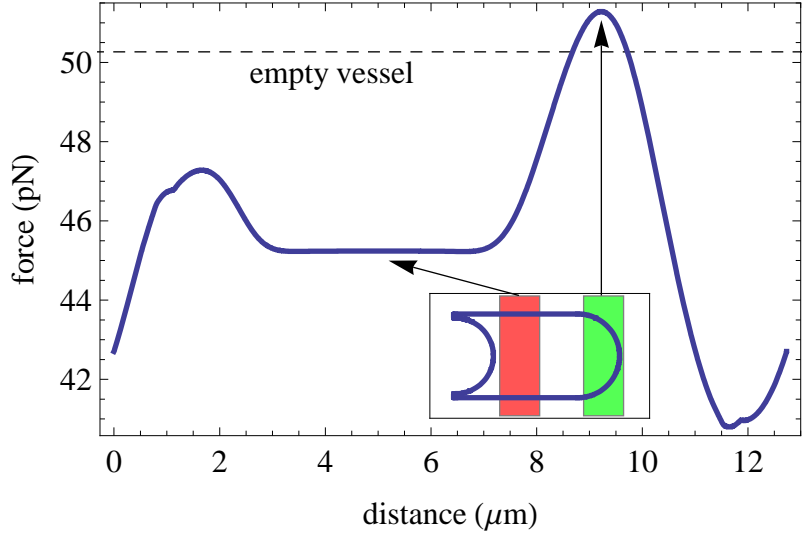}
\end{center}
\caption{Shear force, in piconewtons, on the sensor band as a
function of its position with respect to a passing cell for
$\vCell=1\millimeter/\second$. The dashed line shows the force in
the empty vessel model. The inset shows the position of the sensor
band with respect to the passing cell at two points on the plot,
indicated by the arrows. The total distance shown along the $x$-axis
in the figure corresponds to the cell spacing $L$ of
\tbl{parameters}.}\figlabel{force on band}
\end{figure}

\subsection{Sensor Moving with the Fluid}

For the chemical source on the vessel wall, the fluid flow pushes
the concentration downstream from the source. Thus the sensor
encounters no flux until it reaches and passes the source. As the
sensor moves far downstream of the source it continues to encounter
the chemical, diluted throughout the vessel, that was released by
the source while it was far downstream of the sensor.

\begin{figure}
\begin{center}
\includegraphics*[viewport=0 60 192 100]{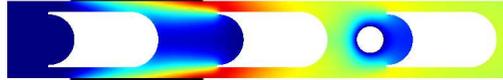}
\end{center}
\caption{\textbf{Sensor moving with the fluid.} Concentration around three cells and the
sensor shortly after it passes the chemical source. Geometry is as
in \fig{spacing} with the addition of the sensor, indicated as the
small white circle between the center and rightmost cells.
Concentration ranges from 0 (dark blue) to $2.5\times
10^{17}\,\molecule/\meter^3$ (red). The black lines along the vessel
wall mark the location of the source at this time, with length
$\sourceLength=10\micron$. The cells, sensor and fluid move from
left to right. Cell speed is $\vCell=1\millimeter/\second$ and
diffusion coefficient is $\Dlarge$.}\figlabel{concentration}
\end{figure}

\fig{concentration} shows one example of how the cells trap the
chemical near the vessel wall for a short distance when the
diffusion constant is relatively small. However, the subsequent
fluid flow and diffusion bring significant flux to the sensor.
\fig{results: sphere} shows how the flux to the sensor varies with
its distance past the source for the low and high diffusion cases.
Low diffusion, i.e., for large molecules, leads to detection further
downstream of the sensor. \fig{results: sphere} also compares the
behavior with the model without the cells. The model without cells
somewhat overestimates the flux to the sensor, with a maximum closer
to the source when diffusion constant is small. These differences
are rather modest, indicating the simpler model is adequate for
estimating sensing behavior.

\begin{figure}
\begin{center}
\includegraphics{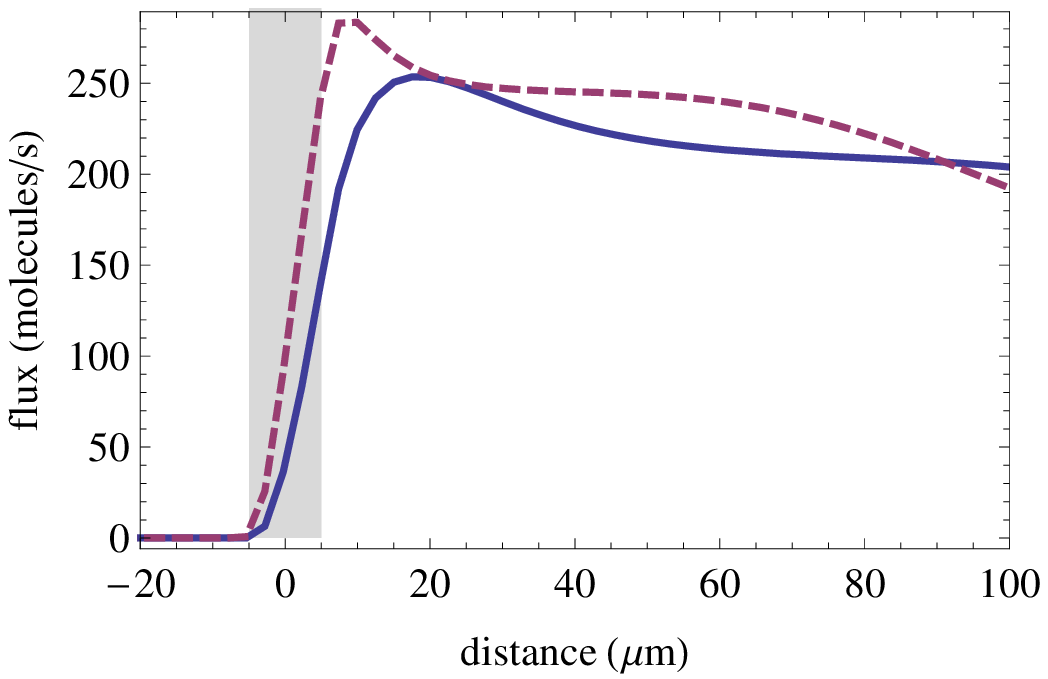}\\
\includegraphics{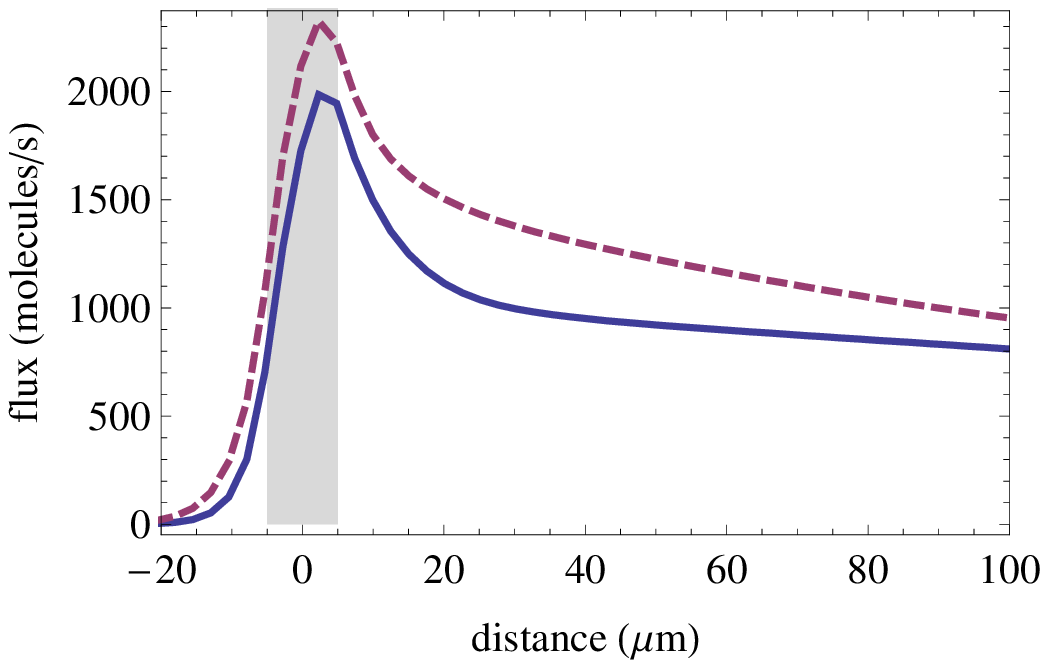}
\end{center}
\caption{Flux to the sensor as a function of distance the center of
the sensor is downstream of the center of the source. The solid and
dashed curves are with and without cells in the vessel,
respectively. Top: large molecules ($\Dlarge$ diffusion
coefficient); bottom: small molecules ($\Dsmall$ diffusion
coefficient). The gray bands show the range of distances where the
center of the sensor overlaps with the source, i.e., where the
sensor could reach the source by moving radially to the vessel
wall.}\figlabel{results: sphere}
\end{figure}

Two quantitative summary measures of sensor performance are the
maximum rate at which the sensor absorbs molecules at it passes the
source, and the number absorbed while the sensor is relatively near
the source. As an example for the latter measure, we give the value
when the center of the sensor is within $20\micron$ of the source,
i.e., for distances in \fig{results: sphere} ranging from $-25$ to
$+25\micron$. The counts while the sensor is near the source
indicate how well the sensor can detect the source while it is still
relatively nearby. \tbl{results: sphere} gives these values for the
different cell speeds and chemical diffusion coefficients, comparing
with and without cells in the vessel. The values differ by only
about $20\%$, indicating models based on empty vessels capture the
main behavior of the sensors. Increasing flow speed reduces the
counts for the sensor, both because the sensor moves through the
high-concentration region near the source more rapidly and also
because the count rate itself decreases. Thus at higher speeds, more
of the chemical moves downstream past the sensor. This is less of an
issue for small molecules, with higher diffusion coefficients.

\begin{table}
\begin{center}
\begin{tabular}{|l|ccc|}
  \hline
  cell speed ($\millimeter/\second$)& $0.2$ & $1$ & $2$ \\
  \hline \multicolumn{4}{|c|}{large molecules}\\
  maximum count rate ($\second^{-1}$) & $900$ & $250$ & $130$ \\
  $\ldots$ without cells & $1000$ & $280$ & $150$ \\
  counts while near source  & $25$ & $5.2$ & $2.7$ \\
  $\ldots$ without cells & $33$ & $6.4$ & $3.3$ \\
  \hline \multicolumn{4}{|c|}{small molecules}\\
  maximum count rate ($\second^{-1}$) & $3000$ & $2000$ & $1500$ \\
  $\ldots$ without cells & $3500$ & $2300$ & $1600$ \\
  counts while near source  & $140$ & $45$ & $27$ \\
  $\ldots$ without cells & $170$ & $58$ & $33$ \\
  \hline
\end{tabular}
\end{center}
\caption{Behavior of the spherical sensor moving with the fluid past
the chemical source on the vessel wall.}\tbllabel{results: sphere}
\end{table}

For comparison, the flux produced by the source is $2\pi R
\sourceLength K \approx 1900\, \molecule/\second$. This is an upper
bound on the \emph{steady-state} flux to the sensor, i.e., if the
sensor captures all the chemical produced by the source. The actual
values of \tbl{results: sphere} are well below this rate, except for
short periods of time with the lower speeds and high diffusion
coefficient.

As another comparison, \eq{absorb} gives the flux to a sphere of
radius $\sensorSize/2$ for a chemical with diffusion coefficient
$\Dlarge$ and concentration far from the sphere equal to the initial
downstream concentration for $\vCell=1\millimeter/\second$ as
$100\,\molecule/\second$.

\section{Discussion}

In summary, for the two sensing scenarios examined, the simpler
model of a vessel without cells gives similar performance estimates
as the model with the cells. Thus neither the change in fluid flow
nor confinement of the chemical significantly alter sensor
performance. Thus the simpler model is sufficient for design studies
of microscopic chemical sensors operating within capillaries. Such
studies are relevant both for diagnostics, where the devices attempt
to detect specific chemicals, and for power generation where the
devices use chemicals in blood plasma as a power source.

On the other hand, the model with the cells identifies additional
aspects of the robots' environment related to the nearby cells. In
particular, the variation in the results show the sensor band on the
vessel wall could estimate the rate of passing cells from variations
in the counts as well as changes in the fluid shear on devices
equipped with flow sensors~\cite{ghosh03}. From such measurements,
the devices could estimate the local hematocrit of the vessel.

The model considers red blood cells of uniform size and spacing.
This approach ignores other components of blood, particularly the
smaller platelets and the much larger white blood cells, which are
much less common than red cells, variation in size and spacing of
the red cells, changes in vessel diameter and effects on the flow at
branching vessels.

As a caveat in interpreting these results, this model is a continuum
approximation to a discrete process of individual molecule
detections. For macroscopic systems, the number of molecules
involved is so large that this approximation accurately represents
the detection rates. However, for microscopic sensors, and
particularly at low concentrations, the number of molecules involved
is relatively small and statistical fluctuations in the actual count
become significant (even assuming the sensor itself makes no errors
in capturing or identifying the molecules that reach it via
diffusion). The continuum approximation gives the average rate of a
Poisson process for the counts, with fluctuations proportional to
the square root of the number. These fluctuations are a significant
source of noise for microscopic sensors, and limit their ability to
discriminate chemical sources of interest from background
concentration from other sources~\cite{hogg06b}.

Moreover, we consider the chemical as arising only from the source
of interest. In practice, there will be background concentrations of
the chemical and the sensing task will need to distinguish high
concentrations over small distances from a lower but pervasive
background concentration. Thus the detection algorithm used by the
sensors must handle statistical fluctuations from the background
concentration occasionally appearing to indicate a high localized
concentration~\cite{hogg06b}. For comparing the models with and
without cells, such background concentration is simple to include in
the plasma and gives a proportionate increase in the detected flux.

Extensions of this model of microscopic devices interacting with
cells in capillaries could address other scenarios. Examples include
chemicals released by the red blood cells (such as oxygen),
branching and changes in size of vessel in which the numerical
method must include changes to cell shape~\cite{mauroy07} and forces
on the cells. The last case is relevant for safety by evaluating
whether the force groups of sensors impose on passing cells could
damage the cells. The model is also relevant for drug delivery tasks
targeted to the bloodstream or its components, and could include
diffusion of these chemicals out of the capillary to surrounding
tissues~\cite{popel89}. Another scenario is the detection of a
transient chemical source rather than a steady-state sources
considered here. In this case, the time history of the transient may
be of interest. The model including cells could determine how cells
in the fluid alter that history as observed by the sensor as an
additional check on the adequacy of the simpler model without cells.

More generally, this work illustrates how models originally
developed to study transport properties in small vessels extend to
engineering design studies of the behaviors and environments of
microscopic robots. By characterizing chemical detection rates and
locations, these models provide constraints on the performance of
individual devices and suggestions for control strategies to enable
robust, rapid task performance of medical tasks by large groups of
robots.

\small
\section*{Acknowledgements}
I thank R. Freitas~Jr for helpful discussions.

\end{document}